\documentclass[letterpaper]{article} 
\usepackage{aaai2026}  
\usepackage{times}  
\usepackage{helvet}  
\usepackage{courier}  
\usepackage[hyphens]{url}  
\usepackage{graphicx} 
\usepackage{natbib}
\usepackage{caption} 
\DeclareCaptionStyle{ruled}{labelfont=normalfont,labelsep=colon,strut=off} 
\usepackage[utf8]{inputenc} 
\usepackage[T1]{fontenc}    
\usepackage{url}            
\usepackage{booktabs}       
\usepackage{amsfonts}       
\usepackage{nicefrac}       
\usepackage{microtype}      
\usepackage{xcolor}         

\usepackage{algorithm}
\usepackage{algpseudocode}
\usepackage{tabularx}

\usepackage{tcolorbox}
\usepackage{amsmath, amssymb, amsthm}
\usepackage{dsfont}
\usepackage{mathtools}
\usepackage{lipsum}
\usepackage{multirow}
\tcbuselibrary{skins}

\usepackage{array}
\usepackage{ragged2e}
\newcolumntype{Y}{>{\RaggedRight\arraybackslash}X}

\newtcolorbox{promptwrapper}{
    enhanced,
    frame hidden,
    boxrule=0pt,
    arc=4pt,  
    outer arc=8pt,  
    colback=blue!5,
    before skip=10pt,
    after skip=10pt,
    overlay={
        \draw[rounded corners=8pt, line width=0.8pt] 
            (frame.south west) rectangle (frame.north east);
    }
}

\newtheorem{proposition}{Proposition}
\newtheorem{definition}{Definition}

\definecolor{darkgreen}{rgb}{0.0, 0.7, 0.0}

\title{Language Model Distillation: A Temporal Difference Imitation Learning Perspective}

\author {
    Zishun Yu\textsuperscript{\rm 1}\equalcontrib,
    Shangzhe Li\textsuperscript{\rm 2}\equalcontrib,
    Xinhua Zhang\textsuperscript{\rm 1}
}
\affiliations {
    \textsuperscript{\rm 1}Department of Computer Science, University of Illinois Chicago\\
    \textsuperscript{\rm 2}Department of Computer Science, University of North Carolina at Chapel Hill\\
    zyu32@uic.edu, shangzhe@unc.edu, zhangx@uic.edu
}

\begin{document}

\maketitle

\begin{abstract}
Large language models have led to significant progress across many NLP tasks, although their massive sizes often incur substantial computational costs. Distillation has become a common practice to compress these large and highly capable models into smaller, more efficient ones. Many existing language model distillation methods can be viewed as behavior cloning from the perspective of imitation learning or inverse reinforcement learning.
This viewpoint has inspired subsequent studies that leverage (inverse) reinforcement learning techniques, including variations of behavior cloning and temporal difference learning methods. 
Rather than proposing yet another specific temporal difference method, we introduce a general framework for temporal difference-based distillation by exploiting the distributional sparsity of the teacher model.
Specifically, it is often observed that language models assign most probability mass to a small subset of tokens. 
Motivated by this observation, we design a temporal difference learning framework that operates on a reduced action space (a subset of vocabulary), and demonstrate how practical algorithms can be derived and the resulting performance improvements.
\end{abstract}

\newcommand{\todo}{\color{blue} TODO}

\newcommand{\E}{\mathbb{E}}

\newcommand{\Scal}{\mathcal{S}}
\newcommand{\Acal}{\mathcal{A}}
\newcommand{\Mcal}{\mathcal{M}}
\newcommand{\Bcal}{\mathcal{B}}

\newcommand{\pe}{\stackrel{p}{=}}

\newcommand{\proj}{\textsc{proj}}

\section{Introduction}

The successes of large language models (LLMs) have brought transformative advancements across many NLP tasks, albeit often at the cost of substantial computational resources due to their massive size. Model distillation is a widely adopted approach to compressing these large, high-capacity models (teachers) into smaller, efficient models (students). This practice has become standard, ranging from proprietary ones~\citep{achiam2023gpt} to open-source model families~\citep{touvron2023llama,grattafiori2024llama, yang2024qwen2}.

The seminal work of knowledge distillation (KD) proposed by \citet{hinton2015distilling} established the foundation for distribution-matching distillation methods, which were later extended to sequence/auto-regressive models~\citep{kim2016sequence}. In distribution-matching methods, at each time step during text generation, 
a student model $\pi: \mathcal{X} \to \Delta(\mathcal{V})$ seeks to approximate the probability distribution, over an vocabulary $\mathcal{V}$, produced by the teacher $\pi^\star$, for example by minimizing a (forward) KL-divergence~\citep{sanh2019distilbert, jiao2019tinybert, wang2020minilm} $\min_{\pi} \mathbb{KL}[ \pi^\star(x_{\leq t}) \Vert\pi(x_{\leq t}) ]$, 
where $x_{\leq t} \in \mathcal{X}$ is a sequence of length $t$, and $\Delta(\Omega)$ denotes a probability simplex over space $\Omega$.

Notably, this sequential distribution matching can be interpreted as behavior cloning (BC)~\citep{pomerleau1991efficient}, within the imitation learning (IL) and inverse reinforcement learning (IRL) literature~\citep{abbeel2004apprenticeship, ng2000algorithms, ziebart2008maximum, ho2016generative}. 
While there are nuanced differences between IL and IRL, we use them interchangeably.

An important distinction between IL and distillation lies in the accessibility of the expert/teacher model $\pi^\star$. In IL settings, $\pi^\star$ is typically not available as a white-box model; instead, it is inferred empirically from expert demonstrations. A common estimate is given by $\hat{\pi}^\star(y \mid x) \propto \mathbb{E}_{(s, a) \sim \rho_\mathbb{D}}[\mathds{1}_{\{(x, y) = (s, a) \}}]$, where $\rho_\mathbb{D}$ is the empirical distribution induced from a dataset $\mathbb{D}$. In contrast, it is reasonable to assume direct access to $\pi^\star$ as a white-box for distillation. This suggests that distillation is an even easier problem to IL, enabling the application of tools and insights from the broader IL literature.

Casting distillation as an IL problem is not new per se; indeed, many existing distillation works~\citep{kim2016sequence, agarwal2024policy, gu2023minillm, wen2023f} perform BC. 
Before reviewing these works through the lens of IL, we highlight two key distinctions that help characterize the distillation setup:
(i) offline vs. online: This refers to whether distillation is performed on a fixed, pre-collected dataset (offline) or involves continuously generating new data during training (online).
(ii) off-policy vs. mixed-policy: A method is considered purely on-policy if it uses only student-generated data for training; otherwise, it is categorized as off-policy. However, purely on-policy distillation is rare in practice—most methods use a mixture of teacher and student data. We refer to this setup as mixed-policy, though it is technically off-policy in the RL literature.

%
%

The pioneer work of sequence-level KD (SeqKD)~\citep{kim2016sequence} is an offline off-policy BC.
Follow up BC style works fall into two main categories: 
(i) online mixed-policy~\citep{agarwal2024policy, gu2023minillm, ko2024distillm}; 
and (ii) offline mixed-policy~\citep{wen2023f}; 
potentially with different choice of divergence measure - forward/reverse KL~\citep{kim2016sequence, gu2023minillm}, Jensen-Shannon divergence~\citep{agarwal2024policy}, and general $f$-divergence~\citep{wen2023f}.

\begin{figure}
    \centering
    \includegraphics[width=0.45\textwidth]{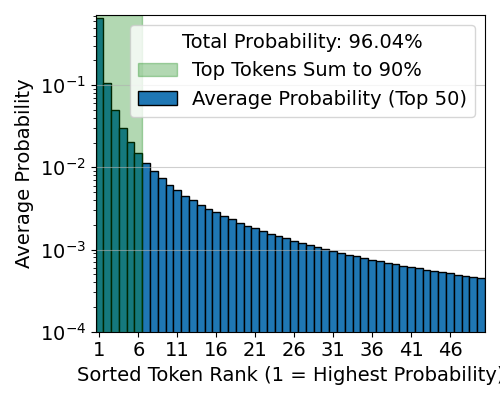}
    \caption{We average the sorted token probabilities across $20$ sequences generated by Qwen-2.5 3B. Top $50$ tokens account for $96\%$ of the total mass, and the top $7$ tokens contribute $\geq 90\%$.}
    \label{fig:teaser}
    \vspace{-2em} 
\end{figure}

However, BC is known to suffer from compounding errors~\citep{ross2011reduction, tu2022sample}, also referred as exposure bias~\citep{ranzato2015sequence} in the context of auto-regressive models.
To address these issues, the IL literature offers a rich toolbox beyond BC, for example generative adversarial training~\citep{ho2016generative}, 
occupancy matching~\citep{syed2008apprenticeship}, 
and more. We refer readers to \citet{osa2018algorithmic} for a comprehensive review.
While not exhaustive, many modern IL methods~\citep{ho2016generative, ziebart2008maximum} involve modeling an inverse reward function and then optimizing a policy with respect to it using RL. These RL algorithms are often based on temporal-difference (TD) learning~\citep{sutton1998reinforcement}, which estimates the long-term impact of an action (or token) given a state (or sub-sequence) to mitigate the compounding errors.


Distillation may potentially benefit from the broader IL toolbox—particularly from TD learning methods. 
Earlier work has already demonstrated the potential of such approaches in smaller auto-regressive models. For instance, \citet{yu2017seqgan} and \citet{wu2021textgail} can be viewed as early applications of \citet{ho2016generative}.
For modern LLMs, \citet{jia2024adversarial} explored TD-style IL methods for distillation, through value moment matching~\citep{abbeel2004apprenticeship, ziebart2008maximum} in particular.
However, applying TD learning to LLMs is generally challenging due to their vast state-action space $\mathcal{V}^T$~\citep{snell2022offline, yu2023mathcal, havrilla2023trlx}.


Instead of directly applying a specific TD-learning-based IL method to distillation, we ask a more fundamental question: is there an exploitable structure unique to the distillation setting?
A natural observation is that, for LLMs, the output distribution over the vocabulary is often sparse—only a small subset of tokens typically receive significant probability mass.
An illustrative example is shown in Figure~\ref{fig:teaser}: For responses we generated, top $50$ tokens account for $96\%$ total mass, and top $7$ tokens contribute $90\%$.
Motivated by this observation, a natural idea is: during TD learning, it may be sufficient to consider only a small subset of candidate actions,  reducing the effective action space from $\mathcal{V}$ to a much smaller set.

\section{Preliminaries}

{\bf MDP notations.} We define the conventional MDP tuple $\mathcal{M} = (\mathcal{S}, \mathcal{A}, \mathcal{P}, r, \gamma)$ as follows.
Let $w_{\leq t} = (w_1, w_2, \dots, w_t) \in \mathcal{W}_t \coloneq \mathcal{V}^t$ be a sequence of tokens. We define a state and an action as $s_t \coloneq s_0 \circ w_{< t} \in \mathcal{S}$ and $a_t \coloneq w_t \in \mathcal{A}$, where $\circ$ denotes concatenation and $s_0$ is a prompt (or a query). We will sometimes use $x$ and $y$ as substitute of $s$ and $a$ whenever required by context, and use the conventional prime notation $s', a'$ to denote $s_{t+1}, a_{t+1}$ when only relative temporal index matters. The transition function $\mathcal{P}:\mathcal{S}\times\mathcal{A} \to \mathcal{S}$ is defined as simple concatenation $s' = \mathcal{P}(s, a) \coloneq s \circ a$; $r:\mathcal{S}\times\mathcal{A}\to \mathbb{R}$ is an unknown reward function; and $\gamma \in [0, 1]$ is a discount factor.

{\bf Policy.} We denote the student policy as $\pi: \mathcal{S} \to \Delta(\mathcal{A})$ (or parameterized $\pi_\theta$) and the teacher model as $\pi^\star: \mathcal{S} \to \Delta(\mathcal{A})$, which may sometimes be referred to as the expert policy using IL terminology.

{\bf Soft value functions.} The goal of soft RL is to maximize the expected return $J(\pi) = \mathbb{E} [ \sum_t \gamma^t \tilde{r}_t \mid \pi,  \mathcal{M}]$. The state-action value function $Q^\pi: \mathcal{S}\times\mathcal{A} \to \mathbb{R}$ and the state value function $V^\pi: \mathcal{S} \to \mathbb{R}$ are essentially recursive definition of the return $J(\pi)$. In particular, we consider soft value functions (entropy-regularized)~\citep{littman1996algorithms, ziebart2008maximum} in this work, which are defined as follows:
\begin{align}
    Q^\pi(s, a) & \coloneq \textstyle\mathbb{E}[\sum_{t=0}^\infty \gamma^t \tilde{r}_t \mid \pi, \mathcal{M}, S_0 = s, A_0 \! = \! a]\nonumber\\
    &= {r}(s, a) + \gamma V^\pi(s'), \\
    V^\pi(s) & \coloneq \mathbb{E}_{a \sim \pi(s)} [Q(s, a) - \log\pi(a\mid s)]
\end{align}
where $s' = s \circ a$, and $\tilde{r}(s, a) \coloneq r(s, a) - \log\pi(a\mid s) $ is the entropy-regularized reward.

In an operator notation, the soft Bellman operator $\mathcal{B}^\pi_r : \mathbb{R}^{\mathcal{S} \times \mathcal{A}} \to \mathbb{R}^{\mathcal{S} \times \mathcal{A}} $ is defined as $(\mathcal{B}^\pi_r Q)(s, a) = r(s, a) + \gamma V^{\pi}(s')$. Note this soft Bellman (evaluation) operator is a contractor~\citep{haarnoja2018soft}, in contrast to the non-contractive soft Bellman improvement operator~\citep{littman1996algorithms}. We will throughout this paper use soft Bellman operator to denote $\mathcal{B}^\pi_r$, the policy evaluation operator. 

{\bf Remark on $\mathcal{B}^\pi_r$.} The contractive property of $\mathcal{B}^\pi_r$ defines a unique soft $Q$-function, for a reward function $r$, as the fixed point of $Q = \mathcal{B}^\pi_r Q$, given the Banach fixed-point theorem~\citep{banach1922operations}. The contraction property of $\mathcal{B}^\pi_r$ later helps us to obtain our results in Section~\ref{sec:method}.

{\bf Occupancy measures.} Let $x$ be a state, the occupancy measure induced from a policy $\pi$ is 
$\rho^\pi_{x}(s, a) \coloneq \mathbb{E}_{\pi} [ \sum^\infty_{t = \tau} \gamma^{t-\tau} \mathds{1}_{\{S_t = s, A_t = a\}} \mid S_\tau = x]$.
Occupancy measures are convenient notation for formulating RL problems, though not a must, we introduce them to make the formulations more succinct. It is convenient as the value function can be written as $V^\pi(s) = \mathbb{E}_{(x, y) \sim \rho_s^\pi} [\tilde{r}(x, y)]$.

{\bf Max-entropy IRL.} With these definitions, we are now ready to introduce our choice of concrete IRL algorithm, IQL~\citep{garg2021iq}, which is an extension of  maximum-entropy IRL~\citep{ziebart2008maximum}. We start with the Generative Adversarial IL (GAIL)~\citep{ho2016generative} objective function:
\begin{equation}
    \min_\pi \max_r L(\pi, r) \coloneq \mathbb{E}_{\rho^\star}[r(s, a)] - \mathbb{E}_{\rho^\pi}[r(s, a)] - \mathcal{H}(\pi) - \psi(r), 
\end{equation}
where $\mathcal{H}(\pi) \coloneq \mathbb{E}_{\rho^\pi}[-\log\pi(a\mid s)]$ and $\psi: \mathbb{R}^{\mathcal{S}\times\mathcal{A} } \to \mathbb{R}$ is a convex reward regularizer.

{\bf Inverse soft Q-learning.} IQL~\citep{garg2021iq} is a recent popular extension of GAIL, which is also the choice of algorithm of our paper. It's objective function is defined as: 
\begin{align}
    \textstyle \max_Q \min_\pi \mathcal{J}(\pi, Q) &\coloneq \mathbb{E}_{\rho^\star}[Q(s, a) - \gamma V^\pi(s')]\nonumber\\
    &- (1-\gamma)\mathbb{E}_{s_0}[V^\pi(s_0)] - \psi(\mathcal{T}^\pi Q), \\
    \textstyle \Leftrightarrow \quad \max_Q \mathcal{J}^\star(Q) &\coloneq \mathbb{E}_{\rho^\star}[\phi(Q(s, a) - \gamma V^Q(s'))]\nonumber\\
    &- (1-\gamma) \mathbb{E}_{s_0} [V^Q(s_0)]
\end{align}
where $\phi$ is a choice of concave regularizer; $\mathcal{T}^\pi: \mathbb{R}^{\mathcal{S}\times\mathcal{A}} \to \mathbb{R}^{\mathcal{S}\times\mathcal{A}}$ is an inverse soft Bellman operator, defined as $(\mathcal{T}^\pi Q)(s, a) = Q(s, a) - \gamma V^\pi(s')$; and $V^Q \coloneq \log\sum_a\exp Q(s, a)$.

The core idea is can be summarized as: 
(i) The original GAIL objective $\min_\pi \max_r  L(\pi, r)$ can be equivalently swapped as $\max_r \min_\pi  L(\pi, r)$ by strong duality, given proper regularization and feasible domains; 
(ii) For a fixed policy $\pi$, the saddle-point problem $\max_r \min_\pi L(\pi, r)$ for a fixed $\pi$ is provably equivalent to $\max_Q \min_\pi \mathcal{J}(\pi, Q)$. 
This equivalence allows direct optimization in the value-function space, eliminating the need to explicitly model the inverse reward $r$;
(iii) For a fixed $Q$, it is well-known that the entropy-regularized policy has a closed-form solution given by $\pi_Q = \exp Q(s, a) / \sum_a \exp Q(s, a)$. 
And Danskin’s theorem~\citep{danskin2012theory} implies the saddle-point optimization $\max_Q \min_\pi \mathcal{J}(\pi, Q)$ can be casted as a simple maximization $\max_Q \mathcal{J}^\star(Q)$.

{\bf Remark on IQL.} 
IQL reparameterizes the saddle-point formulation of IRL, enabling simpler and more stable optimization, and has thus recently gained popularity. 
We refer readers to~\citep{ho2016generative, garg2021iq, al2023ls}, for a detailed derivation, background, and follow-up works. Throughout this paper, we use the deterministic form of IQL since the transition function $\mathcal{P}$ for LLMs is deterministic; the original stochastic formulation can be found in~\citet{garg2021iq}.

\section{Method}\label{sec:method}

\subsection{Notation Setup}

Defining different sort of backup operators with different desired properties has been a significant string of works in the RL literature. We refer to \citet{asadi2017alternative} and \citet{miahiresmax} for detailed discussions. We start with the example of soft Bellman operator used in IQL
\begin{equation}
    \textstyle (\mathcal{B}^\pi_r Q)(s, a) = r(s, a) + \gamma  \mathbb{E}_{a'\sim\pi(s')}[Q(s', a') - \log\pi(a'\mid s')] ]
\end{equation}
The future term $\mathbb{E}_{a'\sim\pi(s')}[Q(s', a') - \log\pi(a'\mid s')]$ is often called target. While evaluating the target, the candidate set of actions is the entire action space $\mathcal{A} = \mathcal{V}$, which is often huge. 

Given the sparsity observation we made in Figure~\ref{fig:teaser}, a natural idea is to only use a subset of candidate actions that have a significant total density mass, say top-$p$ tokens, we define the top-$p$ candidate set:

\begin{definition}[top-$p$ candidate set] Given $\pi^\star$ and a state $s$, the top-$p$ candidate set is defined as
    \begin{align}
        &\textstyle \mathcal{A}^\star_p(s) :=\nonumber\\
        &\{a_i: \textstyle\sum_{i} \pi^\star(a_i \mid s) = p;~ \pi^\star(a_i|s) \geq \pi^\star(a_{j} \mid s)  ~\forall~i, j \} \subseteq \mathcal{A}.
    \end{align}
\end{definition}

where actions are sorted in descending order of probability density (i.e., $\pi^\star(a_i|s)\geq\pi^\star(a_j|s)$ for all $i<j$). Thus, for state $s$, the set $\mathcal{A}^\star_p(s)$ consists of actions with the largest probabilities whose cumulative probability equals $p$, a.k.a. top-$p$ tokens~\citep{holtzman2019curious}. Although it’s uncommon to have a set whose cumulative probability equals exactly $p$, we use this notation for brevity, as the intended meaning is clear from context. 

This top-$p$ set filtered out the actions that have low densities w.r.t. the teacher $\pi^\star$. It is intuitive as the teacher model is a high-capable model and hence the reserved candidate tokens are high-quality.

{\bf Remark on $\Acal^\star_p(s)$.} Note that $\mathcal{A}^\star_p(s)$ is a state-dependent set, which creates notational inconvenience for subsequent analysis. Without loss of generality, we simplify that $\mathcal{A}^\star_p$ is state-independent, it is easy to extend the following results to state-dependent $\mathcal{A}^\star_p(s)$. A simple construction is $\mathcal{A}^\star_p \coloneq \cup_s \mathcal{A}^\star_p(s)$.

Instead proceeding to direct implementation with this top-$p$ set, lets first define a top-$p$ MDP $\mathcal{M}_p$ as a filtered counterpart of the original MDP $\mathcal{M}$. 

\begin{definition}[top-$p$ MDP and $\bar{Q}$] Given a MDP $\mathcal{M} = (\mathcal{S}, \mathcal{A}, \mathbb{P}, r, \gamma)$ and a teacher policy $\pi^\star$, its top-$p$ counterpart is $\mathcal{M}_p = (\mathcal{S}, \mathcal{A}^\star_p, \mathbb{P}, r, \gamma)$. 
To make the notation clearer, we use $Q:\mathcal{S}\times\mathcal{A}\to\mathbb{R}$ and 
{\color{black} $\bar{Q}:\mathcal{S}\times \mathcal{A}_p^\star\to\mathbb{R}$}
to denote $Q$-functions live in $\Mcal$ and $\Mcal_p$, respectively.
\end{definition}
With this definition, we aim to address an important question: {\bf What's the performance of the optimal policy, if exist any, in $\mathcal{M}_p$ compared to the optimal policy in $\mathcal{M}$?  Do we preserve proximal optimality when we use a subset of actions $\mathcal{A}_p^\star$?}

%
Let $\pi^\star$ and $Q^\star$ be the optimal policy and its corresponding soft $Q$-function in $\Mcal$, and let $\pi^\star_p$  and $\bar{Q}_p^\star$ denote their counterpart in $\Mcal_p$. 
As $\Mcal_p$ is also a well-defined MDP, the existence of $\pi^\star, \pi^\star_p, Q^\star, \bar{Q}^\star_p$ follows directly from the fixed point theorem~\citep{banach1922operations}.
Note that we use $\pi^\star$ to refer to both optimal policy and the teacher model, as in IRL it is often assumed that the demonstrator (the teacher) is optimal.

\begin{definition}[top-$p$ projection]\label{def:top-p-projection}
For any policy $\pi$, its top-$p$ counterpart in the top-$p$ MDP $\Mcal_p$ is a state-wise projection $\proj_p$ onto the top-$p$ candidate set $\Acal^\star_p$, for all state $s \in \Scal$.
\begin{equation}
    \proj_p(\pi)(a \mid s) := 
    \begin{cases}
        \pi(a \mid s) / \sum_{a\in \Acal^\star_p} \pi(a \mid s),\;\text{if } a \in \Acal^\star_p \\
        0,\qquad\text{otherwise}
    \end{cases}
\end{equation}
\end{definition}

{\bf Remark on $\proj_p$.} $\proj_p(\pi^\star_p) = \pi^\star_p$  by definition, and $\proj_p(\pi^\star)$ is not necessarily $\pi^\star_p$. However, the projected policy $\proj_p(\pi^\star)$ is handy for our analysis, see later Proposition~\ref{prop:sandwich} and \ref{prop:final-gap}.

With this projection definition, we are now ready to define a top-$p$ version of soft Bellman operator.

\begin{definition}[top-$p$ soft Bellman operator]\label{def:top-p-operator}
For any policy $\pi:\mathcal{S}\times\mathcal{A}\to\Delta(\mathcal{A})$ and reward function $r$, We define a top-$p$ counterpart $\Bcal^\pi_p: \mathbb{R}^{\mathcal{S}\times\mathcal{A}_p^\star} \to \mathbb{R}^{\mathcal{S}\times\mathcal{A}_p^\star}$ to the soft Bellman operator  $\Bcal^\pi_r$ as  
\begin{align}
    (\mathcal{B}^\pi_p \bar{Q}) (s, a) &= 
        r(s, a) + \gamma \E_{a' \sim {\color{blue} \proj_p(\pi)}} [\bar{Q}(s', a')\\
        &- \log{\color{blue} \proj_p(\pi)}(a' \mid s') ].
\end{align}
\end{definition}
The top-$p$ operator $\mathcal{B}^\pi_p$ is defined by projecting $\pi$ onto $\Acal_p^\star$. To characterize the optimality of $\mathcal{B}^\pi_p$, we define the supported $q$-norm to measure the magnitude of vectors w.r.t. the action set of interest $\mathcal{A}_p^\star$:
\begin{definition}[supported $q$-norm]
    For $\Phi: \Scal\times\mathcal{Y}\to \mathbb{R}$, we define the supported $q$-norm by, 
    \begin{equation}
        \textstyle \Vert \Phi\Vert _{q, \mathcal{Y}} := \max\nolimits_s ( \sum\nolimits_{a \in \mathcal{Y}} |\Phi(s, a)|^q )^{1/q}.
    \end{equation}
And slightly abusing notations, for $\Phi: \mathcal{S}\times\Acal\to\mathbb{R}$ and $\bar{\Phi}: \mathcal{S}\times\Acal^\star_p \to\mathbb{R}$, we define
\begin{equation}
    \textstyle \Vert \Phi - \bar\Phi \Vert_{q, \mathcal{A}^\star_p} := \max\nolimits_s ( \sum\nolimits_{a \in \mathcal{A}^\star_p} |\Phi(s, a) - \bar{\Phi}(s, a)|^q )^{1/q}.
\end{equation}
\end{definition}
Note that the supported $q$-norm is not exactly a norm because the definiteness does not hold. However, non-negativity, homogeneity, and especially the triangle inequality do hold.

\subsection{Optimality in $\Mcal_p$}

\begin{definition}[fixed point of $\Bcal^\pi_p$] $\bar{Q}: \Scal \times \Acal_p^\star \to \mathbb{R}$ is a fixed point of $\Bcal^\pi_p$ iff
    \begin{equation}
        \bar{Q} (s, a) = (\Bcal^\pi_p \bar{Q})(s, a) \quad \text{for all $s\in\Scal$ and $a\in\Acal_p^\star$.}
    \end{equation}\label{def:fix-point}
  
\end{definition}

\begin{proposition}[contraction]\label{prop:contraction}
    $\Bcal^\pi_p$ is a contraction in the supported $\infty$-norm.
\end{proposition}

We start with characterizing the gap between $\bar{Q}^{\proj_p{\pi^\star}}$ and $Q^\star$, and then the characterization of $\bar{Q}_p^\star$, the optimal value function in $\Mcal_p$, will follow directly from the sandwich condition~\ref{prop:sandwich}.

\begin{proposition}\label{prop:sandwich-gap}
    Suppose $\kappa(p) := - \frac{\gamma}{1-\gamma} \log p$ and $\bar{Q}^{\proj_p \pi^\star}$ is the fixed point of $\Bcal^{\proj_p \pi^\star}_p$, the sub-optimality $\Vert Q^\star-\bar{Q}^{\proj_p{\pi^\star}} \Vert _{\infty, \Acal^\star_p}\leq \kappa(p) $.
\end{proposition}

Proposition~\ref{prop:sandwich-gap} shows that the sub-optimality between $\bar{Q}^{\proj_p\pi^\star}$ and $Q^\star_p$ is bounded. 
This suggests that the optimal policy $\pi^\star$ in $\Mcal$ projected onto $\mathcal{A}_p^\star$, the action set of $\Mcal_p$, is still a good policy.
While projection is trivial in the tabular case, it is in general difficult to project a parameterized $\pi^\star$ onto the action set $\Acal_p^\star$.
Instead, it is natural to implement an (inverse) RL algorithm in $\Mcal_p$.

However, when implementing conventional IRL algorithms in $\Mcal_p$, these algorithms typically seek to find the optimal policy $\pi^\star_p$ in $\Mcal_p$ rather than the projected $\proj_p \pi^\star$. 
The quantity of interest is the gap $\Vert Q^\star - \bar{Q}^\star_p \Vert_{\infty, \Acal_p^\star}$, which measures how well $\mathcal{M}_p$ approximates the original task $\mathcal{M}$ using the action subset $\mathcal{A}_p^\star$.
This is intuitive: since $\pi_p^\star$ is optimal in $\Mcal_p$, it should perform at least as well as any projected policy, including $\proj_p\pi^\star$.

Formally, we first provide a sandwich condition in Proposition~\ref{prop:sandwich}, and the desired gap, as shown in Proposition~\ref{prop:final-gap}, trivially follows.
\begin{proposition}[sandwich condition]\label{prop:sandwich}
    $ \bar{Q}^{\proj_p{\pi^\star}}(s, a) \leq \bar{Q}^\star_p(s, a) \leq Q^\star(s, a) $, for all $(s, a) \in \Scal\times\Acal^\star_p$.
\end{proposition}

\begin{proposition}[bounded sub-optimality]\label{prop:final-gap} Given a MDP $\Mcal$ and its top-$p$ counterpart $\Mcal_p$, let $Q^\star$ and $\bar{Q}^\star_p$ be their optimal soft $Q$-functions, respectively, we have:
    $\Vert Q^\star-\bar{Q}_p^\star\Vert _{\infty, \Acal^\star_p}\leq \kappa(p) $.
\end{proposition}

The takeaway is: {\bf The optimal policy $\pi^\star_p$ learned in $\Mcal_p$ is provably near-optimal relative to $\pi^\star$ (the teacher), as established by Proposition~\ref{prop:final-gap}}. 
This guarantee means one can deploy \emph{any} IRL algorithm within $\Mcal_p$ and (theoretically) find a policy whose performance closely matches that of $\pi^\star$, even though the action subset $\Acal_p^\star$ (for reasonable choice of $p$) is much smaller than the full action space $\mathcal{A}$, i.e. the raw vocabulary $\mathcal{V}$. This makes TD learning more efficient. 
In the next section, we show how to tailor a concrete IRL algorithm to our top-$p$ MDP $\Mcal$.

\section{Implementation}\label{sec:impl}

We've demonstrated that: Optimal policy $\pi^\star_p$ in $\Mcal_p$ yields a bounded sub-optimality~\ref{prop:final-gap} compared the optimal policy $\pi^\star$ in the original problem $\Mcal$. Technically, any soft IRL algorithm could be applied to $\Mcal_p$, by tailoring its backup operator through top-$p$ projection (definition~\ref{def:top-p-projection}), as the example of top-$p$ soft Bellman operator (definition~\ref{def:top-p-operator}).

We choose IQL~\citep{garg2021iq} as our base IRL algorithm, and show that how to practically implement its top-$p$ counterpart. Recall that the IQL objective is given by:
\begin{align}
    &\textstyle \max_Q \mathcal{J}^\star(Q) \coloneq \mathbb{E}_{\rho^\star}[\phi(Q(s, a) - \gamma V^{\pi_Q}(s'))]\nonumber\\
    &- (1-\gamma) \mathbb{E}_{s_0} [V^{\pi_Q}(s_0)], \textstyle V^{\pi_Q}(s) \coloneq \log\sum_a\exp Q(s, a).\label{eq:iql-obj-original}
\end{align}

Often in (inverse) RL objectives, there are two functions $Q$ and $\pi$ depends on action $a$. Hence to constraint objective~\eqref{eq:iql-obj-original} (or any other objective) to the top-$p$ candidate action set $\mathcal{A}_p^\star$, one should constraint these two functions to the candidate set.

\begin{algorithm}[!t] 
\caption{Bellman Distill}
\begin{algorithmic}\label{algo:pseudo-train}
\State \textbf{Input:} Teacher model $\pi^\star$; Instruction set $\mathcal{D}=\{X\}$;Student model $Q_\theta$; Pre-train dataset $\mathcal{D}^{PT}$.
\State \textbf{Data Generation:} Generate the teacher dataset $\mathcal{D}^\star=\{X,Y^\star\}$ using $\pi^\star$.
\For{each epoch}
    \State Sample mini-batches $\mathcal{D}^\star_{\text{mini}},\mathcal{D}^{PT}_{\text{mini}}$ from datasets.
    \State Compute $\mathcal{F}_p^\star$ according to Definition~\ref{def:top-p-mask} and $Q(s,a)$.
    \State Compute $\pi_Q(a \mid s)$ and $\proj_p \pi_Q $ using Definition~\ref{def:top-p-projection}.
    \State Compute projected value function using $V^{\proj_p\pi_Q}(s)$.
    \State Compute $\mathcal{J}^\star$ in Eq.~\ref{eq:iql-practical} and $\mathcal{J}_{PT}$ for student update.
\EndFor
\end{algorithmic}
\label{alg:pseudo-train}
\end{algorithm}

{\bf $Q$-function masking.}  We only need to update $Q$-values of interests, i.e. $Q(s, a)$ for $a \in \mathcal{A}_p^\star$. 

\begin{definition}[top-$p$ mask]\label{def:top-p-mask}
For any $Q:\mathcal{S}\times\mathcal{A} \to \mathbb{R}$, we define a top-$p$ mask $\mathcal{F}_p^\star: \mathcal{S}\times\mathcal{A} \to \mathcal{S}\times\mathcal{A}$:
\begin{equation}
    (\mathcal{F}_p^\star Q)(s, a) \coloneq 
    \begin{cases}
        Q(s, a), &\text{if~} a \in \mathcal{A}^\star_p \\
        -\infty, &\text{otherwise}
    \end{cases} 
\end{equation}
\end{definition}

Applying $\mathcal{F}_p^\star$ leads to: $\max_Q \mathcal{J}^\star(Q) \coloneq \mathbb{E}_{\rho^\star}[\phi((\mathcal{F}_p^\star Q)(s, a) - \gamma V^{\pi_Q}(s'))] - (1-\gamma) \mathbb{E}_{s_0} [V^{\pi_Q}(s_0)]$.

{\bf Policy projection.} We are also ready to handle the policy $\pi_Q$ through projection. By definition $\pi_Q(a \mid s) = \exp Q(s, a) / \sum_a \exp Q(s, a)$, as a result its projected counterpart is $(\proj_p \pi_Q) = \exp Q(s, a) / \sum_{\mathcal{A}^\star_p}\exp Q(s, a)$ for $a \in \mathcal{A}_p^\star$ otherwise $0$.

Applying both $Q$-function masking and policy projection leads to: 
\begin{align}
    \max_Q \mathcal{J}^\star(Q) &\coloneq \mathbb{E}_{\rho^\star}[\phi((\mathcal{F}_p^\star Q)(s, a) - \gamma V^{\proj_p\pi_Q}(s'))]\nonumber\\
    &- (1-\gamma) \mathbb{E}_{s_0} [V^{\proj_p\pi_Q}(s_0)]
\end{align}

{\bf Further IQL details.} 
$\phi(\cdot)$ is a reward regularizer, we follow~\citet{garg2021iq} to use $\chi^2$, corresponding to $\phi(x) = x-x^2/4\alpha$ for some coefficient $\alpha$. We use $\alpha=0.1$ for our experiments.
In the IQL implementation, the second term in Eq.~\eqref{eq:iql-obj-original} is rewritten equivalently to a TD update through the telescopic identity~\citep{garg2021iq}. This identity states: for any $\pi$, valid occupancy measure $\mu$, and value function $V^\pi$, we have $\mathbb{E}_{(s, a) \sim \mu}\left[V^\pi(s) - \gamma\mathbb{E}_{s'} V^\pi(s')\right] = (1-\gamma) \mathbb{E}_{s_0}[V^\pi(s_0)].$

Let $\mu = \rho^\star$ and $\phi(x) = x-x^2/4\alpha$, the projected IQL objective is thereby given by:
\begin{align}
\label{eq:iql-practical}
    \max_Q \mathcal{J}^\star(Q) &\coloneq \mathbb{E}_{\rho^\star}[\phi((\mathcal{F}_p^\star Q)(s, a) - \gamma V^{\proj_p\pi_Q}(s'))]\nonumber\\
    &- \mathbb{E}_{\rho^\star}\left[V^{\proj_p\pi_Q}(s) - \gamma\mathbb{E}_{s'} V^{\proj_p\pi_Q}(s')\right]
\end{align}
where $V^{\proj_p\pi_Q}(s) = \E_{\proj_p\pi_Q}[Q(s, a) - \log\proj_p\pi_Q]$. Note this is consistent with Eq.~\eqref{eq:iql-obj-original}, since without projection we have $V^{\pi_Q}(s) = \E_{\pi_Q}[Q(s, a)-\log\pi_Q(a\mid s)] = \log\sum_a\exp Q(s, a)$.

Here the expectation w.r.t. $\rho^\star$ implies that data should be sampled from the teacher model $\pi^\star$.

In practice, we parameterize $Q_\theta$ with $\theta$, the parameters of the student model, the student policy $\pi_\theta = \exp Q_\theta(s, a) / \sum_{a\in\mathcal{A}} \exp Q(s, a)$. Therefore, the $Q_\theta$ values are effectively the (constant-shifted) logits of student network, hence it is sufficient to have one model to serve as both $Q_\theta$-function and student policy $\pi_\theta$.

{\bf Further implementation details.} Following~\citet{gu2023minillm}, we maximize a language modeling objective $\mathcal{J}_{PT}=\mathbb{E}_{(s,a)\sim\mathcal{D}^{PT}}\log\pi_Q(a\mid s)$ to retain performance on established NLP benchmarks. We refer to Algorithm~\ref{alg:pseudo-train} and Section 2.3 of \citet{gu2023minillm} for further details. We clip the Q values using a minimum value $Q_\text{min}=-10$ for numerical stability.

\section{Experiments}

In this section, we call our method as Bellman Distill (BD) for brevity.

\begin{figure}
    \centering
    \includegraphics[width=0.45\textwidth]{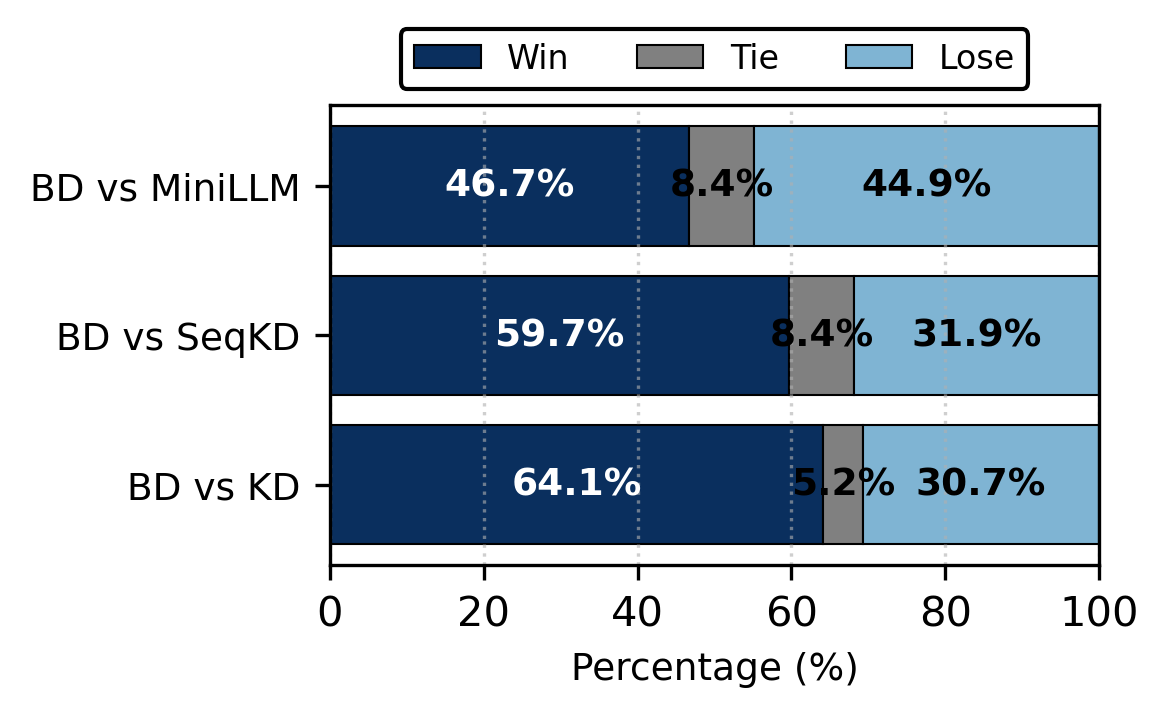}
    \caption{Comparison of win rates against KD, SeqKD, and MiniLLM baselines. We evaluate using GPT-4o-mini \citep{openai2024gpt4omini} as the judging oracle, with Qwen-2.5 (3B) as the teacher model and a smaller (0.5B) model as the student. Results are based on 500 responses per distilled model, generated under the Dolly evaluation setting.}
    \label{fig:win-rate}
    \vspace{-1.5em} 
\end{figure}

\begin{table*}[!t]
    \centering
    \setlength{\tabcolsep}{2.5pt}
    \renewcommand{\arraystretch}{1.1}
    \caption{\textbf{Main results}: The Rouge-L scores, averaged over five random seeds, of our method and baselines across model families. Parameter sizes are listed at the top of each block. Best results per block are in \textbf{bold}. $*$: For GPT-2 class, the baseline results are directly duplicated from \citet{gu2023minillm}.}
    \begin{tabular}{c|ccc|ccc|ccc}
        \toprule 
        \multirow{2}{*}{Method} & \multicolumn{3}{c|}{\textbf{GPT-2$^*$}} & \multicolumn{3}{c|}{\textbf{OPT}} & \multicolumn{3}{c}{\textbf{Qwen-2.5}} \\
        & Dolly & SelfInst & Vicuna & Dolly & SelfInst & Vicuna & Dolly & SelfInst & Vicuna \\
        \midrule
         \multirow{2}{*}{Teacher} & \multicolumn{3}{c|}{1.5B} & \multicolumn{3}{c|}{6.7B} & \multicolumn{3}{c}{3B}\\
        \cmidrule(lr){2-10}
        & 27.6 & 14.3 & 16.3 & 27.6 & 16.4 & 17.8 & 28.8 & 24.1 & 21.0 \\
        \midrule
        Student & \multicolumn{3}{c|}{120M} & \multicolumn{3}{c|}{125M} & \multicolumn{3}{c}{0.5B}\\
        \midrule
        SFT       & 23.2 & 9.9  & 14.3 & 23.2 & 9.9  & 14.3 & 24.5 & 16.5 & 17.6 \\
        KD        & 22.8 & 10.8 & 13.4 & 21.9 & 9.7  & 14.0 & 24.4 & 14.7 & 17.8 \\
        SeqKD     & 22.7 & 10.1 & 14.3 & 22.0 & 10.1 & 13.7 & 24.7 & 15.3 & 17.4 \\
        MiniLLM   & 24.6 & 13.2 & \textbf{16.9} & 23.8 & 10.2 & \textbf{15.3} & 26.7 & 19.1 & 20.5 \\
        BD (Ours) & \textbf{24.7} & \textbf{13.3} & 16.2 & \textbf{25.1} & \textbf{11.4} & 14.8 & \textbf{27.8} & \textbf{19.5} & \textbf{20.7} \\
        \midrule
        Student & \multicolumn{3}{c|}{340M} & \multicolumn{3}{c|}{350M} & \multicolumn{3}{c}{--}\\
        \midrule
        SFT       & 25.5 & 13.0 & 16.0 & 23.6 & 10.6 & 15.5 & --   & --   & -- \\
        KD        & 25.0 & 12.0 & 15.4 & 22.5 & 11.1 & 14.9 & --   & --   & -- \\
        SeqKD     & 25.3 & 12.6 & 16.9 & 23.1 & 11.4 & 14.7 & --   & --   & -- \\
        MiniLLM   & 25.4 & \textbf{15.6} & \textbf{17.7} & 24.3 & 11.5 & 17.9 & --   & --   & -- \\
        BD (Ours) & \textbf{25.8} & 15.4 & 16.5 & \textbf{26.0} & \textbf{12.1} & \textbf{18.1} & --   & --   & -- \\
        \midrule
        Student & \multicolumn{3}{c|}{760M} & \multicolumn{3}{c|}{1.3B} & \multicolumn{3}{c}{--}\\
        \midrule
        SFT       & 25.4 & 12.4 & 16.1 & 26.0 & 11.4 & 15.6 & --   & --   & -- \\
        KD        & 25.9 & 13.4 & 16.9 & 25.5 & 12.0 & 15.4 & --   & --   & -- \\
        SeqKD     & 25.6 & 14.0 & 15.9 & 26.0 & 12.5 & 16.4 & --   & --   & -- \\
        MiniLLM   & \textbf{26.4} & 15.9 & \textbf{18.3} & 26.2 & 13.7 & \textbf{16.7} & --   & --   & -- \\
        BD (Ours) & 26.2 & \textbf{16.1} & 17.3 & \textbf{27.1} & \textbf{15.1} & 16.3 & --   & --   & -- \\
        \bottomrule
    \end{tabular}
    \label{tab:main-results}
\end{table*}

\subsection{Experiment Setup}
We take instruction-following \citep{ouyang2022training} as the conditional text generation task, where models are trained to generate responses according to the instructions. We fine-tune a large model on a dataset consisting of instruction-response pairs as the teacher model. Then, we compare different KD methods by evaluating the student model’s instruction-following performance.

\textbf{Model selections.}
We conduct experiments using three model families: the GPT-2 family \citep{radford2019language}, the OPT family \citep{zhang2022opt}, and the Qwen-2.5 family \citep{yang2024qwen2}. For the GPT-2 family, we use the 1.5B model as the teacher and the 120M, 340M, and 760M models as students. In the OPT family, the 6.7B model serves as the teacher, with the 125M, 350M, and 1.3B models as students. For the Qwen-2.5 family, we use the 3B model as the teacher and the 0.5B model as the student.

\textbf{Training.}
We build our training dataset using \texttt{databricks-dolly-15K} (\url{https://github.com/databrickslabs/dolly/tree/master}), which contains 15,000 human-authored instruction-response pairs. To accommodate model constraints, we remove any samples that exceed the maximum context length. From the remaining data, we randomly select 500 examples for validation and 1,000 for testing, leaving around 12,500 samples for training. For the pretraining dataset $\mathcal{D}^{PT}$, we adopt OpenWebText \citep{Gokaslan2019OpenWeb} for models in the GPT-2 and Qwen-2.5 families, and use the RoBERTa training corpus \citep{liu2019roberta} for OPT models. Following the MiniLLM setup \citep{gu2023minillm}, we select hyperparameters based on Rouge-L \citep{lin-2004-rouge} scores evaluated on the validation set.

As our approach operates in an offline setting, we begin by generating responses from a teacher model using queries from dataset $\mathcal{D}$. Each query may yield multiple responses. These query-response pairs are then used to fine-tune the student models. Additional training details are provided in Appendix B.

{\bf Baselines choices.} Following~\citep{gu2023minillm}, we choose four baselines:
(i) \textbf{SFT} directly fine-tunes the student model on $\mathcal{D}$ with golden responses; (ii) \textbf{KD} \citep{sanh2019distilbert,song2020lightpaff} fine-tunes the student model on $\mathcal{D}$ using the teacher distribution as supervision at each token step,  also known as word-level KD; (iii) \textbf{SeqKD} \citep{kim2016sequence,vicuna2023,alpaca,peng2023instruction,zhou2023lima} fine-tunes the student model on the response sequences generated by the teacher model; (iv) \textbf{MiniLLM} \citep{gu2023minillm} fine-tunes the student model using a policy gradient approach, where the reward is defined by the reverse KL divergence between the output distributions of the teacher and student.

{\bf Evaluation.} Following~\citep{gu2023minillm}, we evaluate the trained models on three instruction-following datasets: (i) \textbf{DollyEval}: The 500-sample test set we split from the \texttt{databricks-dolly-15K} dataset; (ii) \textbf{SelfInst}: \citep{wang2022self} A user-oriented instruction-following set with 252 samples; (iii) \textbf{Vicuna}: \citep{vicuna2023} The 80 challenging questions used in the Vicuna evaluation.

{\bf Metrics.} Following~\citep{gu2023minillm}, we employ two complementary evaluation metrics: (i) \textbf{Rouge-L} \citep{lin-2004-rouge} to assess the precision of generated responses, following prior work demonstrating its effectiveness for large-scale instruction-following evaluation \citep{wang2022super}; (ii) \textbf{Win rate} of our approach evaluated by a GPT-4o-mini \citep{openai2024gpt4omini} oracle against baseline methods to measure generation quality.

A more detailed description on data generation and evaluation can be found in Appendix C.

\begin{table*}[!t]
    \centering
    \setlength{\tabcolsep}{2.5pt}
    \renewcommand{\arraystretch}{1.1}
        \caption{\textbf{Impact of $p$}: We demonstrate the impact of different choices of $p$ across three models from distinct model families. Results are reported using the Rouge-L score and averaged over five random seeds. The performance gains relative to the baseline ($p\,{=}\,1.0$, without top-$p$ mask) are shown in \textcolor{darkgreen}{green}. It is evident that our top-$p$ framework achieves consistent improvements.}
    \resizebox{\textwidth}{!}{
    \begin{tabular}{c|ccc|ccc|ccc}
    \toprule
        \multirow{2}{*}{Models} & \multicolumn{3}{c|}{GPT-2 1.5B $\rightarrow$ 340M} & \multicolumn{3}{c|}{OPT 6.7B $\rightarrow$ 125M} & \multicolumn{3}{c}{Qwen-2.5 3B $\rightarrow$ 0.5B} \\
         & Dolly & SelfInst & Vicuna & Dolly & SelfInst & Vicuna & Dolly & SelfInst & Vicuna\\
        \midrule
        1.0 & 24.9 & 15.2 & 15.7 & 23.6 & 10.5 & 14.2 & 26.6 & 18.1 & 18.6\\
        \midrule
        0.8 & 25.8(\textcolor{darkgreen}{+0.9}) & 15.4(\textcolor{darkgreen}{+0.2}) & 16.5(\textcolor{darkgreen}{+0.8}) & 25.1(\textcolor{darkgreen}{+1.5}) & 11.4(\textcolor{darkgreen}{+0.9}) & 14.8(\textcolor{darkgreen}{+0.6}) & 27.8(\textcolor{darkgreen}{+1.2}) & 19.5(\textcolor{darkgreen}{+1.4}) & 20.7(\textcolor{darkgreen}{+2.1})\\
        0.5 & 25.6(\textcolor{darkgreen}{+0.7}) & 15.5(\textcolor{darkgreen}{+0.3}) & 16.1(\textcolor{darkgreen}{+0.4}) & 24.4(\textcolor{darkgreen}{+0.8}) & 11.0(\textcolor{darkgreen}{+0.5}) & 14.7(\textcolor{darkgreen}{+0.5}) & 27.6(\textcolor{darkgreen}{+1.0}) & 19.2(\textcolor{darkgreen}{+1.1}) & 20.2(\textcolor{darkgreen}{+1.6})\\
        \bottomrule
    \end{tabular}}
    \label{tab:top_p_ablation}
\end{table*}

\begin{table*}[!t]
    \centering
    \caption{\textbf{Training time comparison}: Our approach achieves the optimal validation performance in significantly less training time compared to \citet{gu2023minillm}, thanks to its offline training paradigm.}
    \begin{tabular}{c|ccc}
    \toprule
        Models & GPT-2 1.5B $\rightarrow$ 340M & OPT 6.7B $\rightarrow$ 350M & Qwen-2.5 3B $\rightarrow$ 0.5B \\
    \midrule
        MiniLLM & 11.2h & 12.8h & 10.7h\\
        BD (Ours) & 1.3h & 3.5h & 0.8h \\
    \bottomrule
    \end{tabular}
    \label{tab:training-time}
\end{table*}

\begin{table*}[!t]
    \centering
    \caption{\textbf{$\chi^2$ regularization}: We demonstrate the effectiveness of incorporating $\chi^2$ regularization into the distillation objective. Results reported are the Rouge-L score, averaged over five random seeds.}
    \begin{tabular}{c|ccc|ccc}
    \toprule
        \multirow{2}{*}{Models} & \multicolumn{3}{c|}{GPT-2 1.5B $\rightarrow$ 760M} &  \multicolumn{3}{c}{Qwen-2.5 3B $\rightarrow$ 0.5B} \\
         & Dolly & SelfInst & Vicuna & Dolly & SelfInst & Vicuna\\
        \midrule
        w $\chi^2$ regularization & 26.2 & 16.1 & 17.3 & 27.8 & 19.5 & 20.7 \\
        w/o $\chi^2$ regularization & 25.9 & 15.7 & 17.1 & 27.4 & 19.2 & 20.3 \\
        \bottomrule
    \end{tabular}
    \label{tab:ablation-chi}
\end{table*}

\subsection{Experimental Results}

We present our experimental results in terms of Rouge-L scores for the GPT-2, OPT, and Qwen-2.5 model families in Table~\ref{tab:main-results}. We also demonstrate the win rate results in Figure~\ref{fig:win-rate}. Our method consistently outperforms the KD and SeqKD baselines across all settings. While our approach performs comparably to MiniLLM on the GPT-2 family, it generally surpasses MiniLLM on the OPT and Qwen-2.5 families. Furthermore, the results highlight the scalability of our method across different student model sizes within all three families: as the student model size increases, the Rouge-L score also improves. This demonstrates the strong scalability and generalization ability of our approach across both model sizes and architectures.

\subsection{Impact of $p$}

In this section, we analyze the impact of the top-$p$ value in the top-$p$ distillation setting. We conduct the analysis using GPT-2 340M, OPT 125M, and Qwen-2.5 0.5B models, with results shown in Table~\ref{tab:top_p_ablation} for $p = 0.5$, $0.8$, and $1.0$. The table shows that omitting the top-$p$ mask (i.e., using $p = 1.0$) leads to a notable drop in distillation performance. When $p = 0.5$, performance on OPT 125M is slightly lower than with $p = 0.8$, while the results are comparable for the other two models. Based on this observation, we adopt $p = 0.8$ in most experiments reported in Table~\ref{tab:main-results}, except when using OPT 1.3B as the student model, where $p = 0.5$ yields slightly better results.

\subsection{Training Time}

{\bf Online vs offline training.} In general, the choice between online and offline training is a trade-off between computational cost and performance. While auto-regressive online generation is computationally expensive, offline training allows for the reuse of pre-collected datasets, making it more efficient. However, online training typically yields better final performance~\citep{klein2011batch, lee2019truly, jarrett2020strictly, garg2021iq}. In particular, \citet{wen2023f, ko2024distillm} discussed how the offline setting can accelerate distillation, especially given the high cost of online generation for LLMs. 
We choose offline training due to computational constraints, although the proposed approach is inherently agnostic to the choice of training setting.

{\bf Wall-time comparison.} To demonstrate the computational efficiency of our offline approach, we compare its training time to reach the optimal validation performance with the online method \citet{gu2023minillm}. All experiments are conducted on $4 \times$A40 GPUs. Results are shown in Table~\ref{tab:training-time}.

\subsection{Ablation on the $\chi^2$ Regularization}
We perform an ablation study on the use of $\chi^2$ regularization, as introduced in the implementation section, to assess its impact on distillation performance. Our empirical results show that incorporating $\chi^2$ regularization improves the performance of the distilled student model. The study is conducted on two distillation setups: GPT-2 1.5B to 760M and Qwen-2.5 3B to 0.5B. The results are presented in Table~\ref{tab:ablation-chi}.

\section{Conclusion}
Our work connects language model distillation to imitation learning in large discrete action spaces, where prior methods struggle without action space priors, while distillation benefits from teacher guidance—motivating our top-$p$ TD learning approach. A key limitation is the shared vocabulary requirement between teacher and student for distribution matching. We propose a plug-and-play top-$p$ TD framework that focuses on high-probability tokens, demonstrating empirical gains when integrated with IQL.

\section{Acknowledgment}
We acknowledge support from the NSF ACCESS program for providing computational resources under Award No. CIS250042.
\bibliography{neurips_2025.bib}

@article{hinton2015distilling,
  title={Distilling the knowledge in a neural network},
  author={Hinton, Geoffrey and Vinyals, Oriol and Dean, Jeff},
  journal={arXiv preprint arXiv:1503.02531},
  year={2015}
}

@article{jiao2019tinybert,
  title={Tinybert: Distilling bert for natural language understanding},
  author={Jiao, Xiaoqi and Yin, Yichun and Shang, Lifeng and Jiang, Xin and Chen, Xiao and Li, Linlin and Wang, Fang and Liu, Qun},
  journal={arXiv preprint arXiv:1909.10351},
  year={2019}
}

@article{wang2020minilm,
  title={Minilm: Deep self-attention distillation for task-agnostic compression of pre-trained transformers},
  author={Wang, Wenhui and Wei, Furu and Dong, Li and Bao, Hangbo and Yang, Nan and Zhou, Ming},
  journal={Advances in neural information processing systems},
  volume={33},
  pages={5776--5788},
  year={2020}
}

@article{gu2023minillm,
  title={MiniLLM: Knowledge distillation of large language models},
  author={Gu, Yuxian and Dong, Li and Wei, Furu and Huang, Minlie},
  journal={arXiv preprint arXiv:2306.08543},
  year={2023}
}

@article{jia2024adversarial,
  title={Adversarial Moment-Matching Distillation of Large Language Models},
  author={Jia, Chen},
  journal={arXiv preprint arXiv:2406.02959},
  year={2024}
}

@article{achiam2023gpt,
  title={Gpt-4 technical report},
  author={Achiam, Josh and Adler, Steven and Agarwal, Sandhini and Ahmad, Lama and Akkaya, Ilge and Aleman, Florencia Leoni and Almeida, Diogo and Altenschmidt, Janko and Altman, Sam and Anadkat, Shyamal and others},
  journal={arXiv preprint arXiv:2303.08774},
  year={2023}
}

@article{touvron2023llama,
  title={Llama: Open and efficient foundation language models},
  author={Touvron, Hugo and Lavril, Thibaut and Izacard, Gautier and Martinet, Xavier and Lachaux, Marie-Anne and Lacroix, Timoth{\'e}e and Rozi{\`e}re, Baptiste and Goyal, Naman and Hambro, Eric and Azhar, Faisal and others},
  journal={arXiv preprint arXiv:2302.13971},
  year={2023}
}

@article{grattafiori2024llama,
  title={The llama 3 herd of models},
  author={Grattafiori, Aaron and Dubey, Abhimanyu and Jauhri, Abhinav and Pandey, Abhinav and Kadian, Abhishek and Al-Dahle, Ahmad and Letman, Aiesha and Mathur, Akhil and Schelten, Alan and Vaughan, Alex and others},
  journal={arXiv preprint arXiv:2407.21783},
  year={2024}
}

@inproceedings{kim2016sequence,
  title={Sequence-level knowledge distillation},
  author={Kim, Yoon and Rush, Alexander M},
  booktitle={Proceedings of the 2016 conference on empirical methods in natural language processing},
  pages={1317--1327},
  year={2016}
}

@article{ho2016generative,
  title={Generative adversarial imitation learning},
  author={Ho, Jonathan and Ermon, Stefano},
  journal={Advances in neural information processing systems},
  volume={29},
  year={2016}
}

@article{pomerleau1991efficient,
  title={Efficient training of artificial neural networks for autonomous navigation},
  author={Pomerleau, Dean A},
  journal={Neural computation},
  volume={3},
  number={1},
  pages={88--97},
  year={1991},
  publisher={MIT Press One Rogers Street, Cambridge, MA 02142-1209, USA journals-info~…}
}

@article{osa2018algorithmic,
  title={An algorithmic perspective on imitation learning},
  author={Osa, Takayuki and Pajarinen, Joni and Neumann, Gerhard and Bagnell, J Andrew and Abbeel, Pieter and Peters, Jan and others},
  journal={Foundations and Trends{\textregistered} in Robotics},
  volume={7},
  number={1-2},
  pages={1--179},
  year={2018},
  publisher={Now Publishers, Inc.}
}

@inproceedings{abbeel2004apprenticeship,
  title={Apprenticeship learning via inverse reinforcement learning},
  author={Abbeel, Pieter and Ng, Andrew Y},
  booktitle={Proceedings of the twenty-first international conference on Machine learning},
  pages={1},
  year={2004}
}

@inproceedings{ng2000algorithms,
  title={Algorithms for inverse reinforcement learning.},
  author={Ng, Andrew Y and Russell, Stuart and others},
  booktitle={Icml},
  volume={1},
  number={2},
  pages={2},
  year={2000}
}

@inproceedings{syed2008apprenticeship,
  title={Apprenticeship learning using linear programming},
  author={Syed, Umar and Bowling, Michael and Schapire, Robert E},
  booktitle={Proceedings of the 25th international conference on Machine learning},
  pages={1032--1039},
  year={2008}
}

@inproceedings{ziebart2008maximum,
  title={Maximum entropy inverse reinforcement learning.},
  author={Ziebart, Brian D and Maas, Andrew L and Bagnell, J Andrew and Dey, Anind K and others},
  booktitle={Aaai},
  volume={8},
  pages={1433--1438},
  year={2008},
  organization={Chicago, IL, USA}
}

@inproceedings{agarwal2024policy,
  title={On-policy distillation of language models: Learning from self-generated mistakes},
  author={Agarwal, Rishabh and Vieillard, Nino and Zhou, Yongchao and Stanczyk, Piotr and Garea, Sabela Ramos and Geist, Matthieu and Bachem, Olivier},
  booktitle={The Twelfth International Conference on Learning Representations},
  year={2024}
}

@article{wen2023f,
  title={F-divergence minimization for sequence-level knowledge distillation},
  author={Wen, Yuqiao and Li, Zichao and Du, Wenyu and Mou, Lili},
  journal={arXiv preprint arXiv:2307.15190},
  year={2023}
}

@article{ko2024distillm,
  title={Distillm: Towards streamlined distillation for large language models},
  author={Ko, Jongwoo and Kim, Sungnyun and Chen, Tianyi and Yun, Se-Young},
  journal={arXiv preprint arXiv:2402.03898},
  year={2024}
}

@inproceedings{klein2011batch,
  title={Batch, off-policy and model-free apprenticeship learning},
  author={Klein, Edouard and Geist, Matthieu and Pietquin, Olivier},
  booktitle={European Workshop on Reinforcement Learning},
  pages={285--296},
  year={2011},
  organization={Springer}
}

@article{lee2019truly,
  title={Truly batch apprenticeship learning with deep successor features},
  author={Lee, Donghun and Srinivasan, Srivatsan and Doshi-Velez, Finale},
  journal={arXiv preprint arXiv:1903.10077},
  year={2019}
}

@article{jarrett2020strictly,
  title={Strictly batch imitation learning by energy-based distribution matching},
  author={Jarrett, Daniel and Bica, Ioana and van der Schaar, Mihaela},
  journal={Advances in Neural Information Processing Systems},
  volume={33},
  pages={7354--7365},
  year={2020}
}

@article{garg2021iq,
  title={Iq-learn: Inverse soft-q learning for imitation},
  author={Garg, Divyansh and Chakraborty, Shuvam and Cundy, Chris and Song, Jiaming and Ermon, Stefano},
  journal={Advances in Neural Information Processing Systems},
  volume={34},
  pages={4028--4039},
  year={2021}
}

@article{ranzato2015sequence,
  title={Sequence level training with recurrent neural networks},
  author={Ranzato, Marc'Aurelio and Chopra, Sumit and Auli, Michael and Zaremba, Wojciech},
  journal={arXiv preprint arXiv:1511.06732},
  year={2015}
}

@inproceedings{ross2011reduction,
  title={A reduction of imitation learning and structured prediction to no-regret online learning},
  author={Ross, St{\'e}phane and Gordon, Geoffrey and Bagnell, Drew},
  booktitle={Proceedings of the fourteenth international conference on artificial intelligence and statistics},
  pages={627--635},
  year={2011},
  organization={JMLR Workshop and Conference Proceedings}
}

@inproceedings{tu2022sample,
  title={On the sample complexity of stability constrained imitation learning},
  author={Tu, Stephen and Robey, Alexander and Zhang, Tingnan and Matni, Nikolai},
  booktitle={Learning for Dynamics and Control Conference},
  pages={180--191},
  year={2022},
  organization={PMLR}
}

@book{sutton1998reinforcement,
  title={Reinforcement learning: An introduction},
  author={Sutton, Richard S and Barto, Andrew G and others},
  volume={1},
  number={1},
  year={1998},
  publisher={MIT press Cambridge}
}

@inproceedings{yu2017seqgan,
  title={Seqgan: Sequence generative adversarial nets with policy gradient},
  author={Yu, Lantao and Zhang, Weinan and Wang, Jun and Yu, Yong},
  booktitle={Proceedings of the AAAI conference on artificial intelligence},
  volume={31},
  number={1},
  year={2017}
}

@inproceedings{wu2021textgail,
  title={Textgail: Generative adversarial imitation learning for text generation},
  author={Wu, Qingyang and Li, Lei and Yu, Zhou},
  booktitle={Proceedings of the AAAI Conference on Artificial Intelligence},
  volume={35},
  number={16},
  pages={14067--14075},
  year={2021}
}

@article{snell2022offline,
  title={Offline rl for natural language generation with implicit language q learning},
  author={Snell, Charlie and Kostrikov, Ilya and Su, Yi and Yang, Mengjiao and Levine, Sergey},
  journal={arXiv preprint arXiv:2206.11871},
  year={2022}
}

@article{yu2023mathcal,
  title={$\mathcal{B}$-Coder: Value-Based Deep Reinforcement Learning for Program Synthesis},
  author={Yu, Zishun and Tao, Yunzhe and Chen, Liyu and Sun, Tao and Yang, Hongxia},
  journal={arXiv preprint arXiv:2310.03173},
  year={2023}
}

@inproceedings{havrilla2023trlx,
  title={trlX: A framework for large scale reinforcement learning from human feedback},
  author={Havrilla, Alexander and Zhuravinskyi, Maksym and Phung, Duy and Tiwari, Aman and Tow, Jonathan and Biderman, Stella and Anthony, Quentin and Castricato, Louis},
  booktitle={Proceedings of the 2023 Conference on Empirical Methods in Natural Language Processing},
  pages={8578--8595},
  year={2023}
}

@article{ouyang2022training,
  title={Training language models to follow instructions with human feedback},
  author={Ouyang, Long and Wu, Jeffrey and Jiang, Xu and Almeida, Diogo and Wainwright, Carroll and Mishkin, Pamela and Zhang, Chong and Agarwal, Sandhini and Slama, Katarina and Ray, Alex and others},
  journal={Advances in neural information processing systems},
  volume={35},
  pages={27730--27744},
  year={2022}
}

@article{radford2019language,
title={Language Models are Unsupervised Multitask Learners},
author={Radford, Alec and Wu, Jeff and Child, Rewon and Luan, David and Amodei, Dario and Sutskever, Ilya},
year={2019}
}

@article{zhang2022opt,
  title={Opt: Open pre-trained transformer language models},
  author={Zhang, Susan and Roller, Stephen and Goyal, Naman and Artetxe, Mikel and Chen, Moya and Chen, Shuohui and Dewan, Christopher and Diab, Mona and Li, Xian and Lin, Xi Victoria and others},
  journal={arXiv preprint arXiv:2205.01068},
  year={2022}
}

@article{yang2024qwen2,
  title={Qwen2. 5 technical report},
  author={Yang, An and Yang, Baosong and Zhang, Beichen and Hui, Binyuan and Zheng, Bo and Yu, Bowen and Li, Chengyuan and Liu, Dayiheng and Huang, Fei and Wei, Haoran and others},
  journal={arXiv preprint arXiv:2412.15115},
  year={2024}
}

@misc{Gokaslan2019OpenWeb,  
	title={OpenWebText Corpus},
	author={Aaron Gokaslan and Vanya Cohen},
	howpublished={\url{http://Skylion007.github.io/OpenWebTextCorpus}}, 
	year={2019}
}

@article{liu2019roberta,
  title={Roberta: A robustly optimized bert pretraining approach},
  author={Liu, Yinhan and Ott, Myle and Goyal, Naman and Du, Jingfei and Joshi, Mandar and Chen, Danqi and Levy, Omer and Lewis, Mike and Zettlemoyer, Luke and Stoyanov, Veselin},
  journal={arXiv preprint arXiv:1907.11692},
  year={2019}
}

@inproceedings{lin-2004-rouge,
    title = "{ROUGE}: A Package for Automatic Evaluation of Summaries",
    author = "Lin, Chin-Yew",
    booktitle = "Text Summarization Branches Out",
    month = jul,
    year = "2004",
    address = "Barcelona, Spain",
    publisher = "Association for Computational Linguistics",
    url = "https://aclanthology.org/W04-1013/",
    pages = "74--81"
}

@article{sanh2019distilbert,
  title={DistilBERT, a distilled version of BERT: smaller, faster, cheaper and lighter},
  author={Sanh, Victor and Debut, Lysandre and Chaumond, Julien and Wolf, Thomas},
  journal={arXiv preprint arXiv:1910.01108},
  year={2019}
}

@article{song2020lightpaff,
  title={LightPAFF: A two-stage distillation framework for pre-training and fine-tuning},
  author={Song, Kaitao and Sun, Hao and Tan, Xu and Qin, Tao and Lu, Jianfeng and Liu, Hongzhi and Liu, Tie-Yan},
  journal={arXiv preprint arXiv:2004.12817},
  year={2020}
}

@misc{vicuna2023,
    title = {Vicuna: An Open-Source Chatbot Impressing GPT-4 with 90\%* ChatGPT Quality},
    url = {https://lmsys.org/blog/2023-03-30-vicuna/},
    author = {Chiang, Wei-Lin and Li, Zhuohan and Lin, Zi and Sheng, Ying and Wu, Zhanghao and Zhang, Hao and Zheng, Lianmin and Zhuang, Siyuan and Zhuang, Yonghao and Gonzalez, Joseph E. and Stoica, Ion and Xing, Eric P.},
    month = {March},
    year = {2023}
}

@misc{alpaca,
  author = {Rohan Taori and Ishaan Gulrajani and Tianyi Zhang and Yann Dubois and Xuechen Li and Carlos Guestrin and Percy Liang and Tatsunori B. Hashimoto },
  title = {Stanford Alpaca: An Instruction-following LLaMA model},
  year = {2023},
  publisher = {GitHub},
  journal = {GitHub repository},
}

@article{peng2023instruction,
  title={Instruction tuning with gpt-4},
  author={Peng, Baolin and Li, Chunyuan and He, Pengcheng and Galley, Michel and Gao, Jianfeng},
  journal={arXiv preprint arXiv:2304.03277},
  year={2023}
}

@article{zhou2023lima,
  title={Lima: Less is more for alignment},
  author={Zhou, Chunting and Liu, Pengfei and Xu, Puxin and Iyer, Srinivasan and Sun, Jiao and Mao, Yuning and Ma, Xuezhe and Efrat, Avia and Yu, Ping and Yu, Lili and others},
  journal={Advances in Neural Information Processing Systems},
  volume={36},
  pages={55006--55021},
  year={2023}
}

@article{wang2022self,
  title={Self-instruct: Aligning language models with self-generated instructions},
  author={Wang, Yizhong and Kordi, Yeganeh and Mishra, Swaroop and Liu, Alisa and Smith, Noah A and Khashabi, Daniel and Hajishirzi, Hannaneh},
  journal={arXiv preprint arXiv:2212.10560},
  year={2022}
}

@article{wang2022super,
  title={Super-naturalinstructions: Generalization via declarative instructions on 1600+ nlp tasks},
  author={Wang, Yizhong and Mishra, Swaroop and Alipoormolabashi, Pegah and Kordi, Yeganeh and Mirzaei, Amirreza and Arunkumar, Anjana and Ashok, Arjun and Dhanasekaran, Arut Selvan and Naik, Atharva and Stap, David and others},
  journal={arXiv preprint arXiv:2204.07705},
  year={2022}
}

@book{littman1996algorithms,
  title={Algorithms for sequential decision-making},
  author={Littman, Michael Lederman},
  year={1996},
  publisher={Brown University}
}

@inproceedings{haarnoja2018soft,
  title={Soft actor-critic: Off-policy maximum entropy deep reinforcement learning with a stochastic actor},
  author={Haarnoja, Tuomas and Zhou, Aurick and Abbeel, Pieter and Levine, Sergey},
  booktitle={International conference on machine learning},
  pages={1861--1870},
  year={2018},
  organization={Pmlr}
}

@article{banach1922operations,
  title={Sur les op{\'e}rations dans les ensembles abstraits et leur application aux {\'e}quations int{\'e}grales},
  author={Banach, Stefan},
  journal={Fundamenta mathematicae},
  volume={3},
  number={1},
  pages={133--181},
  year={1922},
  publisher={Polska Akademia Nauk. Instytut Matematyczny PAN}
}

@book{danskin2012theory,
  title={The theory of max-min and its application to weapons allocation problems},
  author={Danskin, John M},
  volume={5},
  year={2012},
  publisher={Springer Science \& Business Media}
}

@article{al2023ls,
  title={Ls-iq: Implicit reward regularization for inverse reinforcement learning},
  author={Al-Hafez, Firas and Tateo, Davide and Arenz, Oleg and Zhao, Guoping and Peters, Jan},
  journal={arXiv preprint arXiv:2303.00599},
  year={2023}
}

@inproceedings{asadi2017alternative,
  title={An alternative softmax operator for reinforcement learning},
  author={Asadi, Kavosh and Littman, Michael L},
  booktitle={International Conference on Machine Learning},
  pages={243--252},
  year={2017},
  organization={PMLR}
}

@article{miahiresmax,
  title={Resmax: An Alternative Soft-Greedy Operator for Reinforcement Learning},
    year={2024},
  author={Miahi, Erfan and MacQueen, Revan and Ayoub, Alex and Masoumzadeh, Abbas and White, Martha},
  journal={Transactions on Machine Learning Research},

}

@article{holtzman2019curious,
  title={The curious case of neural text degeneration},
  author={Holtzman, Ari and Buys, Jan and Du, Li and Forbes, Maxwell and Choi, Yejin},
  journal={arXiv preprint arXiv:1904.09751},
  year={2019}
}

@misc{openai2024gpt4omini,
  author       = {OpenAI},
  title        = {GPT-4o mini: advancing cost-efficient intelligence},
  year         = {2024},
  url          = {https://openai.com/index/gpt-4o-mini-advancing-cost-efficient-intelligence/},
}




\onecolumn
\section{Technical Appendix}
\appendix
\section{Omitted Proofs}

\subsection{Proposition 1}

\begin{proof}
For any $\bar{Q}:\mathcal{S}\times\mathcal{A}_p^\star \to \mathbb{R}$ and any policy $\pi:\mathcal{S}\to\Delta(\mathcal{A})$
\begin{align*}
    &\Vert \mathcal{B}^\pi_p \bar{Q}_1 - \mathcal{B}^\pi_p \bar{Q}_2 \Vert_{\infty, \mathcal{A}_p^\star} \\ 
    \intertext{by definition of supported $\infty$ norm and $\mathcal{B}^\pi_p$, and $r(s, a)$ cancels, }
    &\leq \max_{s, a} \gamma | 
    \E_{a' \sim \proj_p(\pi)}[\bar{Q}_1(s', a') - \log\proj_p(\pi)(a'| s')]
    -\E_{a' \sim \proj_p(\pi)}[\bar{Q}_2(s', a') - \log\proj_p(\pi)(a'| s')]
    |   \\
    &= \max_{s, a} \gamma | 
    \E_{a' \sim \proj_p(\pi)}[\bar{Q}_1(s', a')]
    -\E_{a' \sim \proj_p(\pi)}[\bar{Q}_2(s', a')]
    |   \\
    &\leq \gamma \max_{s'} \max_{a'\in\mathcal{A}_p^\star} |\bar{Q}_1(s', a') - \bar{Q}_2(s', a')| \eqcolon \Vert  \bar{Q}_1 -  \bar{Q}_2 \Vert_{\infty, \mathcal{A}_p^\star}
\end{align*}
Hence $\mathcal{B}^\pi_p$ is contraction in $\Vert \cdot \Vert_{\infty, \mathcal{A}_p^\star}$.
\end{proof}

\subsection{Proposition 2}

\begin{proof}
    For brevity, we drop the norm subscripts $\infty$ and $\mathcal{A}_p^\star$, and we use $\hat{Q}_p^\star$ and $\hat{\Bcal}^\star_p$ to denote $\bar{Q}^{\proj_p \pi^\star}$ and $\Bcal^{\proj_p \pi^\star}_p$, respectively.
\begin{align*}
    & \underbrace{\Vert Q^\star - \hat{Q}^{\star}_p \Vert _{\infty, \Acal_p^\star}}_{=: \delta} = \Vert \mathcal{B}^\star Q^\star - \hat{\mathcal{B}}^\star_p \hat{Q}^{\star}_p \Vert  \leq \Vert \hat{\mathcal{B}}^\star_p Q^\star - \hat{\mathcal{B}}^\star_p \hat{Q}^{\star}_p \Vert  + \Vert \mathcal{B}^\star Q^\star-\hat{\mathcal{B}}^\star_p Q^\star \Vert  \\  
    \intertext{since $\hat{\Bcal}^\star_p$ is a contractor,} \\
    & \leq \gamma \delta + \gamma \max_{s, a\in\mathcal{A}_p^\star} \left| \E_{\pi^\star}\left[(Q^\star-\log\pi^\star)(y|x) \right] -  \E_{\proj_p\pi^\star}\left[(Q^\star-\log\proj_p\pi^\star)(y|x) \right] \right| \\
    & = \gamma \delta +  \gamma \max_{s, a\in\mathcal{A}_p^\star} \left| \sum_{y \in \Acal} \pi^\star(y|x) \left(Q^\star - \log\pi^\star \right)(y|x) - 
    \sum_{y \in  \Acal_p^\star} \frac{1}{p}\pi^\star(y|x)\left(Q^\star - \log\frac{1}{p}\pi^\star \right)(y|x) \right| \\ 
    &= \gamma \delta +  \gamma \max_{s, a\in\mathcal{A}_p^\star} \left| \sum_{y \in \Acal} \pi^\star(y|x) \left(Q^\star - \log\pi^\star \right)(y|x) - 
    \frac{1}{p} \sum_{y \in \Acal_p^\star} \pi^\star(y|x)\left(Q^\star - \log \pi^\star \right)(y|x) - \log p \right| \\
    \intertext{given $\pi^\star(a|s) = \exp{Q^\star(s, a)} / Z(s)$, where $Z(s):=\sum_\Acal \exp Q^\star(s, a)$.}
    &= \gamma \delta +  \gamma \max_{s, a\in\mathcal{A}_p^\star} \left| \sum_{y \in \Acal} \pi^\star(y|x) \log Z(x) - 
    \frac{1}{p} \sum_{y \in \Acal_p^\star} \pi^\star(y|x) \log Z(x) - \log p \right| \\
    \intertext{since $\sum_{y\in\mathcal{A}_p^\star}\pi(y|x) = p$}
    & = \gamma\delta  - \gamma \log p
\end{align*}
Rearranging concludes the proof.
\end{proof}

\subsection{Proposition 3}

\begin{proof}
(i) $Q^\star(s, a) \geq \bar{Q}^\star_p(s, a)$ follows trivially from the fact that $Q^\star$ is the optimal value function of $\Mcal$. By definition, $Q^\star(s, a)\geq Q^\pi(s, a)$ for all $s\in \Scal, a \in \Acal$. 

Hence $Q^\star(s, a)\geq Q^{\pi^\star_p}(s, a) \eqcolon \bar{Q}^\star(s, a)$ holds for all $s\in \Scal, a \in \Acal_p^\star$ as $\Acal_p^\star \subseteq \Acal$; 

(ii) Similarly, $\bar{Q}^\star_p$ is the optimal soft $Q$-function in $\Mcal_p$. We have $\bar{Q}_p^\star(s, a) \geq \bar{Q}^\pi_p(s, a)$ for all $s\in\mathcal{S}, a\in\mathcal{A}_p^\star$. This holds for $\proj_p \pi$ as well, thereby $\bar{Q}^\star(s, a) \geq \bar{Q}^{\proj_p \pi^\star}(s, a)$ for all $s\in\mathcal{S}, a\in\mathcal{A}_p^\star$.

(i) and (ii) conclude the proof.

\end{proof}

\subsection{Proposition 4}

\begin{proof}
    Immediately follows from Proposition 2 and the sandwich condition Proposition 3.
\end{proof}

\section{Training Details}
\label{sec:training-details}

Across all settings, our BD distillation process uses a batch size of 64 and a learning rate of 5e-6. We set \( p = 0.8 \) for all experiments, except for the OPT 6.7B to 1.3B distillation, where we use \( p = 0.5 \). We use the regularization strength $\alpha=0.1$ and the discount factor $\gamma=0.99$ for all settings. Following the setup of \citet{gu2023minillm}, we also adopt a two-phase training strategy for our approach:
\begin{itemize}
    \item \textbf{Phase 1:} We fine-tune the student model on the instruction-response training set $\mathcal{D}$ to obtain a strong initialization for subsequent offline BD training. This fine-tuning is performed for 3 epochs using the optimal learning rate and batch size from the corresponding SFT baselines. Note that, unlike the SFT baseline, we select the checkpoint with the lowest validation loss during this phase, rather than the one with the highest Rouge-L score.
    \item \textbf{Phase 2:} We then further fine-tune the initialized student model using our distillation algorithm 1 on the teacher-generated dataset \(\mathcal{D}^\star\). The student model is trained for 3 epochs, and we select the checkpoint with the best validation Rouge-L score. All experiments are conducted on a machine with \(4 \times\) A40 GPUs.

\end{itemize}

\section{Data Generation and Evaluation Details}
\label{sec:eval-details}

During data generation and evaluation, we sample from each model using a temperature of 1, conditioned on a given query from the corresponding dataset. For evaluation, we use five random seeds—$[10, 20, 30, 40, 50]$—to perform sampling. During data generation, we sample 8 responses from the teacher model for each query in the prompt dataset to construct the teacher offline dataset $\mathcal{D}_T$. To convert instruction-response pairs into complete sentences, we apply the prompt wrapper shown in Figure~\ref{fig:prompt-wrapper}.

\begin{figure}[h]
    \centering
    \begin{promptwrapper}
    \texttt{Below is an instruction that describes a task.}\\
    \texttt{Write a response that appropriately completes the request.}
    
    \medskip
    
    \texttt{\#\#\# Instruction:}\\
    \{\texttt{instruction}\}
    
    \medskip
    
    \texttt{\#\#\# Input:}\\
    \{\texttt{input}\}
    
    \medskip
    
    \texttt{\#\#\# Response:}\\
    \{\texttt{response}\}
    
    \end{promptwrapper}
    \caption{Prompt wrapper for evaluation and data generation.}
    \label{fig:prompt-wrapper}
\end{figure}

For the win rate evaluation, we use the prompt wrapper shown in Figure~\ref{fig:prompt-wrapper-gpt4} to format the instruction along with one response from our method and one from a baseline method into a single sentence, which is then evaluated by the GPT-4o-mini \citep{openai2024gpt4omini} oracle. To mitigate potential bias from the ordering of options, we conduct each comparison twice—once with our method as option A and the baseline as option B, and once with the roles reversed. We then average the win rates from both evaluations.

\begin{figure}[h]
    \centering
    \begin{promptwrapper}
    \texttt{You are an impartial evaluator.}\\
    \texttt{Given a question instruction, optional input context, and two candidate answers (A and B), decide which answer is better or if they tie. Respond with exactly one token: A, B, or Tie.}
    
    \medskip
    
    \texttt{\#\#\# QUESTION INSTRUCTION:}\\
    \{\texttt{instruction}\}
    
    \medskip
    
    \texttt{\#\#\# ADDITIONAL CONTEXT:}\\
    \{\texttt{input}\}
    
    \medskip
    
    \texttt{\#\#\# ANSWER A:}\\
    \{\texttt{response A}\}

    \medskip
    
    \texttt{\#\#\# ANSWER B:}\\
    \{\texttt{response B}\}

    \medskip

    \texttt{Which answer is better?}
    
    \end{promptwrapper}
    \caption{Prompt wrapper for win rate evaluation using GPT-4o-mini \citep{openai2024gpt4omini}.}
    \label{fig:prompt-wrapper-gpt4}
\end{figure}

\section{Generated Examples}
We present several instruction-following examples in the Vicuna setting for the distilled Qwen-2.5 model using our approach. These examples are shown in Table~\ref{tab:response-cases}.
\begin{table}[ht]
\centering
\small
\caption{Examples of instruction-following responses from a distilled Qwen-2.5 0.5B model using our distillation approach.}
\begin{tabular}{@{}p{0.4\textwidth}|p{0.55\textwidth}@{}}
\toprule
\textbf{Instruction} & \textbf{Generated Response} \\
\midrule
How can you determine if a restaurant is popular among locals or mainly attracts tourists, and why might this information be useful? (category: \textit{common-sense}) &
There are several metrics you can look at to determine how popular a restaurant is. One metric is the number of reviews and the type of review (in print or digital) and length of the reviews. You can look at these to see what type of reviews are coming from local residents versus tourists. There are also metrics to look at from a business perspective like average waiting time, and availability of menu changes. All of these can be useful to understand how popular a restaurant is and why that information is important. \\
\midrule
What are the primary factors that influence consumer behavior? (category: \textit{generic}) & There are many factors that influence consumer behavior. 

These include:

    - Economic factors such as income, savings and budget
    
    - Social factors such as values, culture and fashion
    
    - Psychological factors such as mood and emotion
    
    - Consumer preferences and trends.\\
\midrule
In a world where automation is becoming increasingly prevalent, is it more important to prioritize job creation or technological progress? (category: \textit{common-sense})&
In a world where job creation is becoming increasingly difficult to do, prioritizing technological progress is more important. Automation is a tool that can be used to help create new jobs, but it is also a tool that can be use to replace jobs that are currently existent.\\
\bottomrule
\end{tabular}
\label{tab:response-cases}
\end{table}

\end{document}